\documentclass[11pt,a4paper]{article}

\usepackage[utf8]{inputenc}
\usepackage[T1]{fontenc}
\usepackage{lmodern}

\usepackage[
    a4paper,
    left=2.5cm,
    right=2.5cm,
    top=2.5cm,
    bottom=2.5cm
]{geometry}

\usepackage{microtype}
\usepackage{setspace}
\usepackage{amsmath}
\usepackage{amssymb}
\usepackage{bm}

\usepackage{graphicx}
\usepackage{booktabs}
\usepackage{multirow}
\usepackage{array}
\usepackage{longtable}
\usepackage{pdflscape}
\usepackage{float}

\usepackage[numbers,sort&compress]{natbib}

\usepackage[
    colorlinks=true,
    linkcolor=blue,
    citecolor=blue,
    urlcolor=blue
]{hyperref}

\usepackage{titlesec}

\titleformat{\section}
{\large\bfseries}
{\thesection.}{0.5em}{}

\titleformat{\subsection}
{\normalsize\bfseries}
{\thesubsection.}{0.5em}{}

\DeclareUnicodeCharacter{2013}{--}
\DeclareUnicodeCharacter{2014}{---}
\DeclareUnicodeCharacter{2212}{-}
\DeclareUnicodeCharacter{00D7}{\ensuremath{\times}}
\DeclareUnicodeCharacter{00A0}{ }

\usepackage{authblk}

\title{
\textbf{
Physically Aware Radiomics Without Interpolation:
Disentangling Voxel Geometry and Signal Modification in CT and MRI
}
}

\author[1]{David Corral Fontecha}
\author[2,3,4,5]{Juan Miranda Bautista}
\author[2,3,4,5]{Pablo Menendez Fernández-Miranda}
\author[6]{Sergio Rubio-Martín}
\author[7]{Lara Lloret Iglesias}
\author[5,8]{Jose A. Vega}

\affil[1]{Department of Radiology, Complejo Asistencial Universitario de León, León, Spain}
\affil[2]{Department of Radiology, Hospital Universitario Rey Juan Carlos, Móstoles, Madrid, Spain}
\affil[3]{Health Research Institute of the Jiménez Díaz Foundation, IIS-FJD, Madrid, Spain}
\affil[4]{Department of Physical Therapy, Occupational Therapy, Rehabilitation and Physical Medicine, Rey Juan Carlos University, Madrid, Spain}
\affil[5]{Department of Morphology and Cell Biology, Grupo SINPOS, Universidad de Oviedo, Oviedo, Spain}
\affil[6]{Department of Electrical, Systems and Automation Engineering, Universidad de León, León, Spain}
\affil[7]{Advanced Computing and e-Science Group, IFCA-CSIC, Santander, Spain}
\affil[8]{Facultad de Ciencias de la Salud, Universidad Autónoma de Chile, Providencia-Santiago, Chile}
\date{}

\begin{document}

\maketitle

\begin{abstract}

\textbf{Objective:}
Radiomic texture features are computed from voxel-index neighborhoods, implicitly assuming isotropic spatial relationships. In anisotropic medical images, this may confound physical voxel geometry with interpolation-induced signal modification. We developed and evaluated a voxel-spacing--aware radiomic framework that incorporates physical voxel geometry into neighborhood-dependent feature computation without image resampling.

\textbf{Approach:}
A modified PyRadiomics backend was implemented to account for voxel geometry during texture computation while preserving the native image signal. Four geometric configurations were compared: native non-resampled extraction (NR), isotropic resampling (RS), voxel-spacing--aware extraction (VS), and fake-isotropic preprocessing (FK), in which spacing metadata were overwritten without modifying the image array. Experiments were performed on 685 pulmonary nodules from LIDC-IDRI CT and 209 breast MRI cases from I-SPY2. A total of 196 radiomic descriptors were analyzed. Robustness was assessed using ICC(A,1), ICC(C,1), within-subject coefficient of variation, relative median absolute deviation, Friedman testing, ensemble feature selection, machine learning models, a multilayer perceptron baseline, and external validation.

\textbf{Main results:}
VS showed near-native agreement with NR in both modalities. Median ICC(A,1) for NR--VS was 0.9976 in CT and 0.9984 in MRI, whereas RS produced lower agreement and larger feature deviations. NR and VS yielded identical discretization parameters, while RS reduced gradient-derived bin widths in both CT and MRI. Geometry sensitivity was feature-family dependent, with gradient-derived and neighborhood-sensitive texture descriptors showing the greatest preprocessing dependence. FK showed intermediate behavior, indicating that spacing metadata alone can influence radiomic features. In external validation, VS preserved predictive performance comparable to NR in CT, whereas MRI showed greater variability across preprocessing strategies and classifiers.

\textbf{Significance:}
Voxel geometry influences radiomic feature behavior. Voxel-spacing--aware extraction separates geometric modeling from interpolation-induced signal modification while preserving the native image signal. These findings support physically aware radiomic extraction as a coherent alternative to isotropic resampling for quantitative imaging biomarker analysis in anisotropic CT and MRI.

\end{abstract}

\section{Introduction}

Radiomics extracts quantitative descriptors from medical images to characterize tissue phenotype and support diagnosis, prognosis and treatment response prediction. Although radiomics has shown considerable promise for precision oncology in computed tomography (CT) and magnetic resonance imaging (MRI), clinical translation remains limited by poor reproducibility across acquisition and preprocessing conditions~\cite{aerts2014decoding,lambin2017radiomics,van2017computational,mayerhoefer2020introduction}. 

Despite this potential, radiomics remains highly sensitive to non-biological sources of variability, which limits reproducibility and hinders clinical translation. Variations in acquisition protocols, reconstruction parameters and scanner hardware can significantly alter radiomic feature values, even when the underlying biology is unchanged~\cite{mackin2015measuring,berenguer2018radiomics,zwanenburg2020image}. Among these factors, voxel geometry—and particularly voxel anisotropy—represents a fundamental yet often underappreciated source of variability.

In routine clinical imaging, voxel spacing is frequently anisotropic. CT acquisitions typically combine high in-plane resolution with substantially thicker axial slices, whereas MRI anisotropy additionally reflects trade-offs involving signal-to-noise ratio, acquisition time and spatial coverage. Such anisotropy directly affects neighborhood-dependent texture descriptors including GLCM, GLRLM and related feature families by altering the physical interpretation of spatial relationships and directional connectivity~\cite{zhao2016reproducibility,shafiq2017intrinsic}. 

The most widely adopted strategy to address voxel anisotropy is isotropic resampling, typically performed by interpolating images to a fixed voxel spacing (e.g., $1\times1\times1$ mm). Although this enforces geometric consistency, interpolation inherently modifies the original image signal, introduces artificial spatial correlations and may suppress biologically relevant heterogeneity~\cite{larue2017influence,whybra2019assessing}. Conversely, direct extraction from native images preserves the original voxel array but implicitly treats voxel neighborhoods as index-based rather than physical-space relationships, an assumption that becomes problematic in anisotropic acquisitions. 

To address these limitations, voxel-spacing--aware radiomic extraction has been proposed as a physically grounded alternative. Instead of modifying the image through interpolation, spacing-aware methods incorporate the true voxel dimensions into neighborhood definitions and distance-dependent computations, enabling feature extraction directly in physical space while avoiding interpolation of the original image signal~\cite{cattell2019robustness,shafiq2018voxel,park2021robustness}. 

However, several key challenges remain. First, most studies addressing voxel geometry in radiomics have focused primarily on CT, where intensities are physically standardized but voxel geometry is highly variable. It remains unclear whether geometry-aware extraction generalizes to MRI, where signal intensity is non-standardized and acquisition-dependent. Moreover, multiple preprocessing operations are frequently combined within radiomic pipelines, making it difficult to isolate the independent contribution of voxel geometry handling to downstream feature behavior and model performance~\cite{zwanenburg2019radiomics}.

Most conventional radiomic workflows treat voxel neighborhoods as index-based rather than physical-space relationships, potentially compromising the interpretation of texture descriptors in anisotropic images. To address this limitation, we evaluated a voxel-spacing--aware (VS) radiomic framework designed to incorporate true physical voxel geometry directly into neighborhood-dependent feature computation without interpolation of the native image signal.

\begin{figure}[htbp]
\centering
\includegraphics[width=\textwidth]{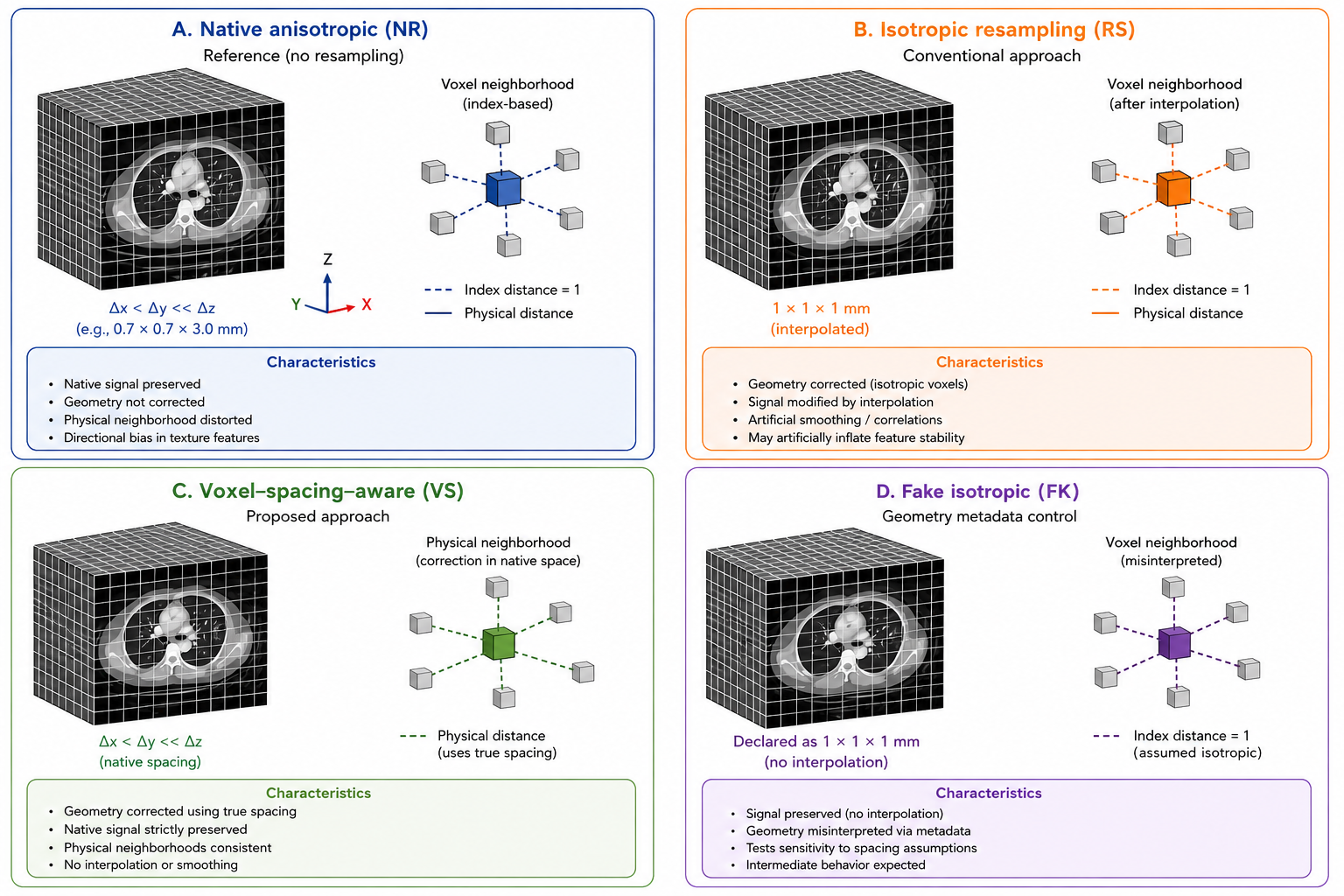}
\caption{
Conceptual overview of the four geometric preprocessing configurations evaluated in this study. Native anisotropic extraction (NR) preserves the original voxel array but treats local neighborhoods in index space, which may distort their physical interpretation. Isotropic resampling (RS) enforces an isotropic voxel grid through interpolation, thereby correcting geometry while modifying the original image signal. Voxel-spacing--aware extraction (VS) preserves the native voxel array while incorporating true physical voxel spacing into neighborhood-based computations, thereby separating geometric correction from interpolation-induced signal modification. Fake isotropic preprocessing (FK) preserves the original voxel array but overwrites spacing metadata, serving as a control to isolate the effect of geometric reinterpretation alone.
}
\label{fig:conceptual_geometry_interpolation}
\end{figure}

Using a controlled modality-aware experimental design across CT and MRI, we compared four geometric preprocessing configurations: native non-resampled extraction (NR), isotropic resampling (RS), voxel-spacing--aware extraction (VS) and a fake-isotropic control (FK) designed to isolate geometric metadata effects independently from interpolation (Figure~\ref{fig:conceptual_geometry_interpolation}). All methods were implemented within a modified PyRadiomics framework while preserving compatibility with standard extraction workflows. Feature robustness, variability, ensemble feature-selection behavior and downstream predictive performance were evaluated using statistical analysis, machine learning models and a deep learning baseline.

We hypothesized that explicitly accounting for physical voxel geometry during feature extraction—without interpolation—would preserve native radiomic behavior while improving the physical interpretability and methodological coherence of radiomic analysis in anisotropic medical imaging.

\section{Materials and Methods}

\subsection{Datasets and Preprocessing}

The study was conducted using two independent public imaging cohorts representing complementary clinical and physical imaging scenarios in computed tomography (CT) and magnetic resonance imaging (MRI).

The primary CT cohort was derived from the publicly available Lung Image Database Consortium and Image Database Resource Initiative (LIDC-IDRI) dataset~\cite{armato2011lung}. Pulmonary nodules between 6 and 30 mm in diameter were selected following established clinical recommendations and prior radiomics studies~\cite{macmahon2017guidelines}. Nodules segmented by at least three radiologists were retained, while lesions with ambiguous malignancy scores (score 3) were excluded to reduce label uncertainty. One representative lesion per scan was selected, yielding a final curated cohort of 685 pulmonary nodules.

Only CT volumes with valid spatial metadata and compatible axial orientation were included. Histograms of slice thickness and voxel spacing confirmed substantial inter-study anisotropy and acquisition heterogeneity across the cohort.

Segmentation masks were reconstructed using probabilistic fusion of multiple annotations with majority voting and soft thresholding. Final masks were spatially aligned to the reference geometry and stored in NIfTI format.

Benign nodules were generally smaller than malignant lesions, as expected clinically. Because geometry-related descriptors are intrinsically influenced by lesion size, primary comparative analyses focused mainly on intensity- and texture-derived radiomic features.

The secondary cohort consisted of breast MRI studies from the publicly available I-SPY2 trial available through The Cancer Imaging Archive (TCIA)~\cite{li2022ispy2}. Dynamic contrast-enhanced T1-weighted sequences acquired approximately 3 minutes after contrast administration were selected as the reference imaging volumes for analysis.

Tumor masks were reconstructed from the provided annotation volumes and mapped back to the corresponding reference image geometry. All segmentations were manually reviewed and minor morphological corrections were applied when necessary.

After quality control and preprocessing, a final cohort of 209 MRI cases with known pathological complete response status was obtained for radiomic analysis.

MRI analyses were conducted independently from CT due to fundamental differences in intensity physics and normalization requirements. Nevertheless, the same geometric preprocessing and radiomic extraction framework was applied consistently across both modalities. For MRI, intensity normalization was additionally performed using the native normalization procedures implemented within the PyRadiomics pipeline to account for the absence of standardized physical intensity units.

To evaluate downstream generalizability, independent private external validation cohorts were additionally included. For CT, a histopathologically confirmed private cohort of 46 pulmonary nodules with benign and malignant diagnoses was used. For MRI, an independent private cohort of 63 breast cancer cases was employed. All external validation segmentations were manually performed or corrected by an experienced radiologist.

\subsection{Voxel-Spacing–Aware Implementation}

The PyRadiomics framework was extended to support voxel-spacing–aware texture extraction directly in native image space without interpolation. The implementation was integrated across the Python frontend, C wrapper layer and computational backend while preserving full backward compatibility with standard PyRadiomics behavior when spacing-aware extraction was disabled.

Rather than applying a single correction strategy to all texture families, spacing information was incorporated according to the mathematical structure of each descriptor family. GLCM descriptors were computed using standard directional co-occurrence matrices followed by anisotropy-aware angular aggregation. NGTDM descriptors incorporated anisotropy-aware weighted neighborhood averaging. In contrast, GLRLM, GLDM and GLSZM descriptors were computed on a finite-volume zero-order-hold representation of the native voxel grid, preserving original gray-level values while expressing run length, dependence size and zone size on an isotropic physical lattice.

Shape features already operate in physical coordinates within the original PyRadiomics framework and therefore required only minor consistency adjustments. First-order descriptors remained unchanged except for quantities already dependent on voxel volume.

All modifications were implemented as optional extensions activated only in voxel-spacing–aware mode. The complete implementation has been integrated into a functional modified PyRadiomics framework and is currently being prepared for public release and upstream contribution.

\subsection{Physical-Space Neighborhood Modeling}

Conventional radiomic extraction defines spatial relationships using voxel-index offsets. In anisotropic images, however, identical index displacements may correspond to substantially different physical distances. To account for this discrepancy, voxel spacing information was incorporated into texture computation while preserving the native voxel array and avoiding interpolation-induced intensity modification.

For direction-dependent descriptors such as GLCM, feature values were first computed independently for each directional matrix and subsequently combined using anisotropy-aware angular weighting. For NGTDM, local neighborhood means were computed using spacing-aware weighted averaging, allowing neighboring voxels to contribute according to their relative physical separation.

For GLRLM, GLDM and GLSZM, voxel anisotropy was handled using a finite-volume representation of the native image grid. Each voxel was expanded into isotropic subcells using a zero-order-hold strategy in which all subcells inherited the gray level of the original voxel. No interpolation, smoothing or intensity mixing was performed. Texture matrices were then computed on this expanded isotropic lattice, allowing run length, dependence size and zone size to be expressed in physically consistent units while preserving the original image signal.

Importantly, the proposed framework preserves native image intensities and voxel topology while introducing physical-space awareness only where mathematically justified. The present work focuses on the biological and methodological consequences of spacing-aware extraction. A complete technical description of the implementation and software architecture is reported separately.

\subsection{Radiomic Feature Extraction Framework}

Radiomic features were extracted using a modified version of the PyRadiomics framework~\cite{van2017computational}, selected for its widespread adoption and conceptual compatibility with Image Biomarker Standardisation Initiative (IBSI) feature definitions~\cite{zwanenburg2020image}. In the default extraction mode, the framework preserves standard PyRadiomics/IBSI-compatible behavior. The voxel-spacing--aware mode implemented in this study should therefore be interpreted as an optional physical-space extension of selected neighborhood-dependent texture computations, rather than as a modification of the standard IBSI feature definitions themselves.

All processing was performed in Python using standard scientific and medical imaging libraries, including SimpleITK, NumPy, Pandas and Scikit-learn.

A total of 98 radiomic features were extracted per image type from both original and gradient magnitude images, yielding 196 radiomic descriptors in total, including first-order statistics, shape descriptors and texture features derived from GLCM, GLRLM, GLSZM, GLDM and NGTDM matrices.

Feature extraction was intentionally restricted to original and gradient-filtered images to reduce the confounding influence of higher-order transformations and preserve direct physical interpretability of voxel geometry effects.

\subsection{Geometric Handling Configurations}

The primary experimental factor corresponded to voxel geometry handling during radiomic extraction.

Four geometric configurations were evaluated:

\begin{itemize}

\item \textbf{No Resampling (NR):}
Direct feature extraction from native-space images without interpolation or geometric correction.

\item \textbf{Isotropic Resampling (RS):}
Images resampled to isotropic spacing ($1\times1\times1$ mm) using B-spline interpolation for images and nearest-neighbor interpolation for masks.

\item \textbf{Voxel-Spacing--Aware Extraction (VS):}
Feature extraction performed directly in native image space while explicitly incorporating physical voxel dimensions into neighborhood-dependent texture computations.

\item \textbf{Fake Isotropic (FK):}
Negative geometric control in which voxel spacing metadata were artificially overwritten to isotropic values (1 × 1 × 1 mm) while preserving the original voxel array without interpolation.

\end{itemize}

The FK configuration was included to isolate the influence of geometric metadata independently from interpolation-induced signal modification.

\subsection{Gray-Level Discretization}

Gray-level discretization was performed using a fixed bin width strategy following IBSI recommendations and prior radiomics reproducibility studies~\cite{larue2017influence}.

Bin width values were estimated independently for Original and Gradient-filtered images using a target discretization of approximately 50 gray-level bins across lesions. To reduce sensitivity to extreme outliers, robust percentile-based estimates of lesion intensity ranges were used when necessary.

The resulting discretization parameters were subsequently fixed and applied consistently across all geometric preprocessing configurations to ensure that observed differences primarily reflected voxel geometry handling rather than discretization variability.

\subsection{Statistical Analysis}

Statistical analyses were performed feature-wise across all geometric preprocessing configurations described above to evaluate the effect of voxel geometry handling on radiomic robustness, stability and biological discrimination.

Feature reproducibility across geometric configurations was assessed using intraclass correlation coefficients. Both single-measure absolute-agreement ICC, denoted ICC(A,1) and single-measure consistency ICC, denoted ICC(C,1), were computed. ICC(A,1) was used to quantify whether feature values remained numerically comparable across preprocessing strategies, whereas ICC(C,1) assessed preservation of relative subject ranking. These metrics were computed globally across NR, RS, VS and FK and also separately within benign and malignant groups.

To complement ICC analysis, within-subject variability across geometric configurations was quantified using the median within-subject coefficient of variation (wCV) and relative median absolute deviation (rMAD). These metrics were also computed overall and separately within benign and malignant lesions to reduce confounding by biological class differences.

Pairwise robustness analyses were additionally performed between preprocessing strategies, including NR--VS, NR--RS, NR--FK, RS--VS, FK--VS and FK--RS. For each pair and each feature, ICC(A,1), ICC(C,1), mean absolute difference, median absolute difference and signed bias were computed. These pairwise comparisons were used to assess whether VS preserved native feature behavior more closely than RS and whether FK behaved as an appropriate negative geometric control.

Global differences among preprocessing configurations were assessed using the Friedman test for paired repeated measurements. When relevant, pairwise differences between preprocessing strategies were evaluated using Wilcoxon signed-rank tests. Multiple comparisons were corrected using the Benjamini--Hochberg false discovery rate procedure~\cite{benjamini1995controlling}.

Biological discrimination between benign and malignant lesions was evaluated independently within each geometric configuration. Normality and homoscedasticity were assessed using Shapiro--Wilk and Levene tests, respectively. When both assumptions were satisfied, independent-samples t-tests were used and effect size was reported as Cohen's $d$. Otherwise, Mann--Whitney U tests were applied and effect size was reported as rank-biserial correlation. P-values were corrected separately within each preprocessing configuration using Benjamini--Hochberg false discovery rate correction.

\subsection{Ensemble Feature Analysis}

An ensemble-based feature analysis was performed independently for NR, RS, VS and FK to identify radiomic descriptors consistently selected across preprocessing configurations.

Only numeric radiomic features were considered. Diagnostic variables, acquisition metadata, voxel spacing descriptors and non-feature columns were excluded. Shape features were omitted from the primary ensemble analysis to reduce geometry-dominated confounding and focus on intensity- and texture-derived descriptors.

A 5-fold stratified cross-validation framework was used. Within each fold, preprocessing and feature selection were learned exclusively from the training partition to avoid information leakage. Missing values were imputed using training-set medians and features were standardized before selection.

Feature ranking combined complementary statistical and machine learning approaches, including Mann--Whitney U testing with Benjamini--Hochberg correction, recursive feature elimination with logistic regression, Spearman correlation, chi-square selection, absolute logistic regression coefficients, Random Forest importance and LightGBM importance estimation.

Each selector retained up to 30 features per fold. Features receiving at least four selector votes within a fold were considered selected. Cross-fold stability was defined as the proportion of folds in which a feature satisfied this criterion and features selected in at least 40\% of folds were labeled as stable ensemble descriptors.

A secondary ultraconservative sensitivity analysis was additionally performed using stricter consensus thresholds requiring at least five selector votes and selection in at least 60\% of folds.

Final feature rankings were ordered according to stability status, selection frequency and selector agreement to evaluate preprocessing-dependent feature-selection behavior across geometric handling strategies.

\subsection{Machine Learning Modeling}

Machine learning analyses were performed to evaluate the effect of geometric preprocessing on downstream predictive performance.

All preprocessing, feature selection and hyperparameter optimization procedures were learned exclusively from training data and subsequently applied as frozen transformations to validation and test partitions.

Logistic Regression and XGBoost were selected as representative linear and non-linear classifiers.

Predictive signatures were generated using a train-only ensemble feature-selection strategy independent from the interpretative stability analysis described above. In the primary setting, a common neutral signature was constructed from training-only rankings shared across preprocessing configurations, enabling direct comparison between NR, RS, VS and FK under identical feature spaces while minimizing geometry-dependent feature-selection bias.

Additional analyses included preprocessing-specific ensemble signatures and unrestricted full-feature baselines without dimensionality reduction.

Features selected by at least four independent methods were retained, up to a maximum of 30 descriptors per model. Hyperparameter optimization was performed using 5-fold stratified cross-validation within the training set. Final models were subsequently refitted on the complete training partition and evaluated on a held-out internal test set.

A compact tabular multilayer perceptron (MLP) was additionally evaluated as an exploratory deep learning baseline to assess whether geometry-dependent preprocessing effects persisted beyond conventional machine learning classifiers.

\subsection{External Validation}

To assess cross-dataset generalizability, the best-performing models derived from the primary cohort were additionally evaluated on independent private datasets acquired under different clinical and technical conditions.

\begin{figure}[htbp]
\centering
\includegraphics[width=\textwidth]{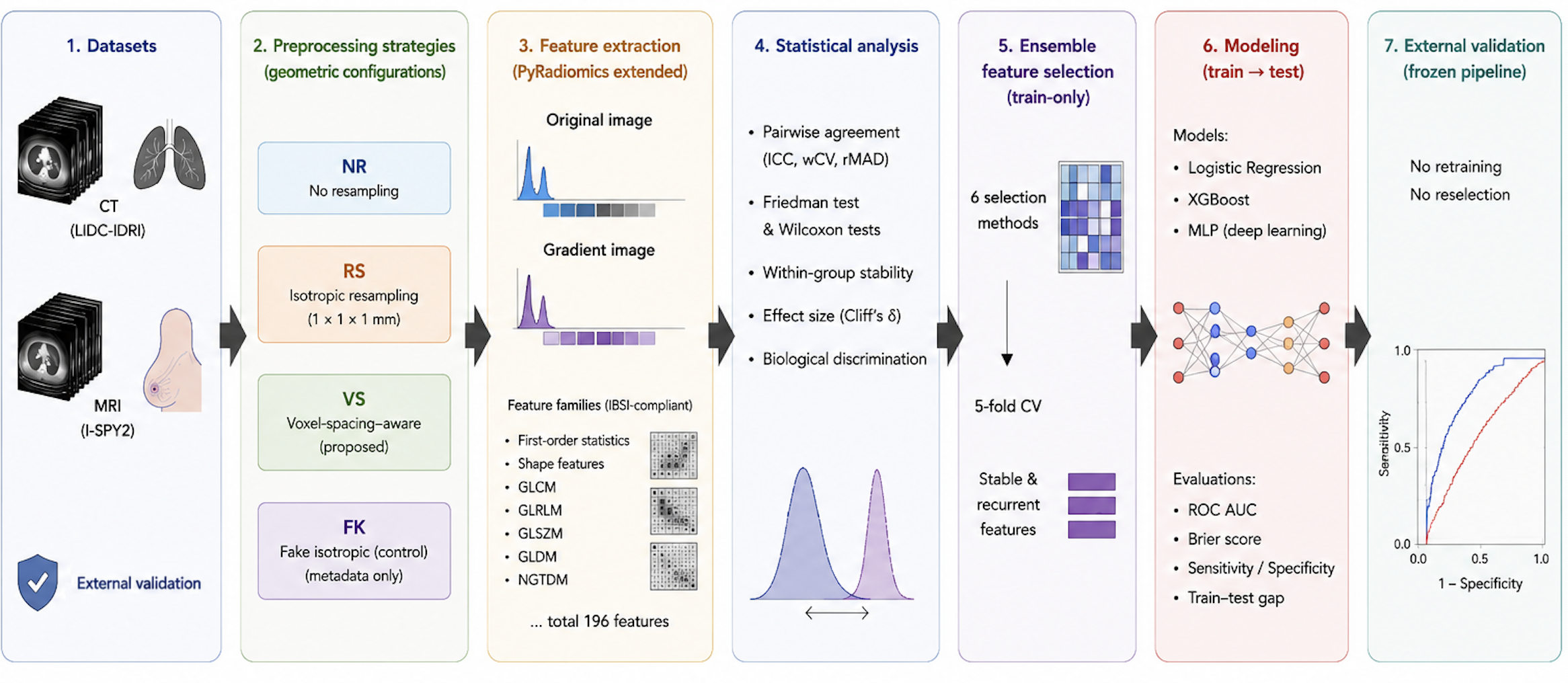}
\caption{
Overview of the geometry-focused multimodal radiomics workflow. CT and MRI cohorts were processed using four geometric configurations: native non-resampled extraction (NR), isotropic resampling (RS), voxel-spacing--aware extraction (VS) and fake isotropic preprocessing (FK). Radiomic features were extracted using an extended PyRadiomics backend from original and Gradient-filtered images. Subsequent analyses included feature robustness statistics, ensemble feature selection, leakage-free machine learning and deep learning modeling and frozen-pipeline external validation.
}
\label{fig:study_workflow}
\end{figure}

No model retraining or feature reselection was performed on the external cohorts. All preprocessing transformations, feature signatures and trained model parameters were transferred directly from the original training pipeline. The overall workflow is summarized in Figure~\ref{fig:study_workflow}.

\section{Results}

\subsection{Bin Width and Geometry-Dependent Discretization}

Gray-level discretization demonstrated a consistent geometry-dependent pattern across CT and MRI.

NR and VS produced identical optimal bin width values for both Original and Gradient-filtered images in the two modalities, indicating preservation of the native intensity distribution despite the use of physically redefined neighborhood relationships. In contrast, isotropic resampling consistently reduced optimal bin width values, particularly for Gradient-filtered representations. FK demonstrated intermediate behavior, showing that modification of spacing metadata alone partially altered Gradient-derived discretization despite preservation of the original voxel array.

Differences between preprocessing strategies were more pronounced for Gradient-filtered images than for Original representations (Table~\ref{tab:binwidth_summary}).

\begin{table}[ht]
\centering
\caption{Optimal bin width values obtained for Original and Gradient-filtered images across geometric preprocessing configurations in CT and MRI.}
\label{tab:binwidth_summary}
\begin{tabular}{llcc}
\hline
\textbf{Modality} & \textbf{Method} & \textbf{Original} & \textbf{Gradient} \\
\hline
CT  & NR & 21.92 & 14.63 \\
CT  & VS & 21.92 & 14.63 \\
CT  & RS & 21.78 & 11.90 \\
CT  & FK & 21.92 & 11.45 \\
\hline
MRI & NR & 99.78 & 62.37 \\
MRI & VS & 99.78 & 62.37 \\
MRI & RS & 95.29 & 48.67 \\
MRI & FK & 99.78 & 46.41 \\
\hline
\end{tabular}
\end{table}

\begin{figure}[htbp]
\centering
\includegraphics[width=\textwidth]{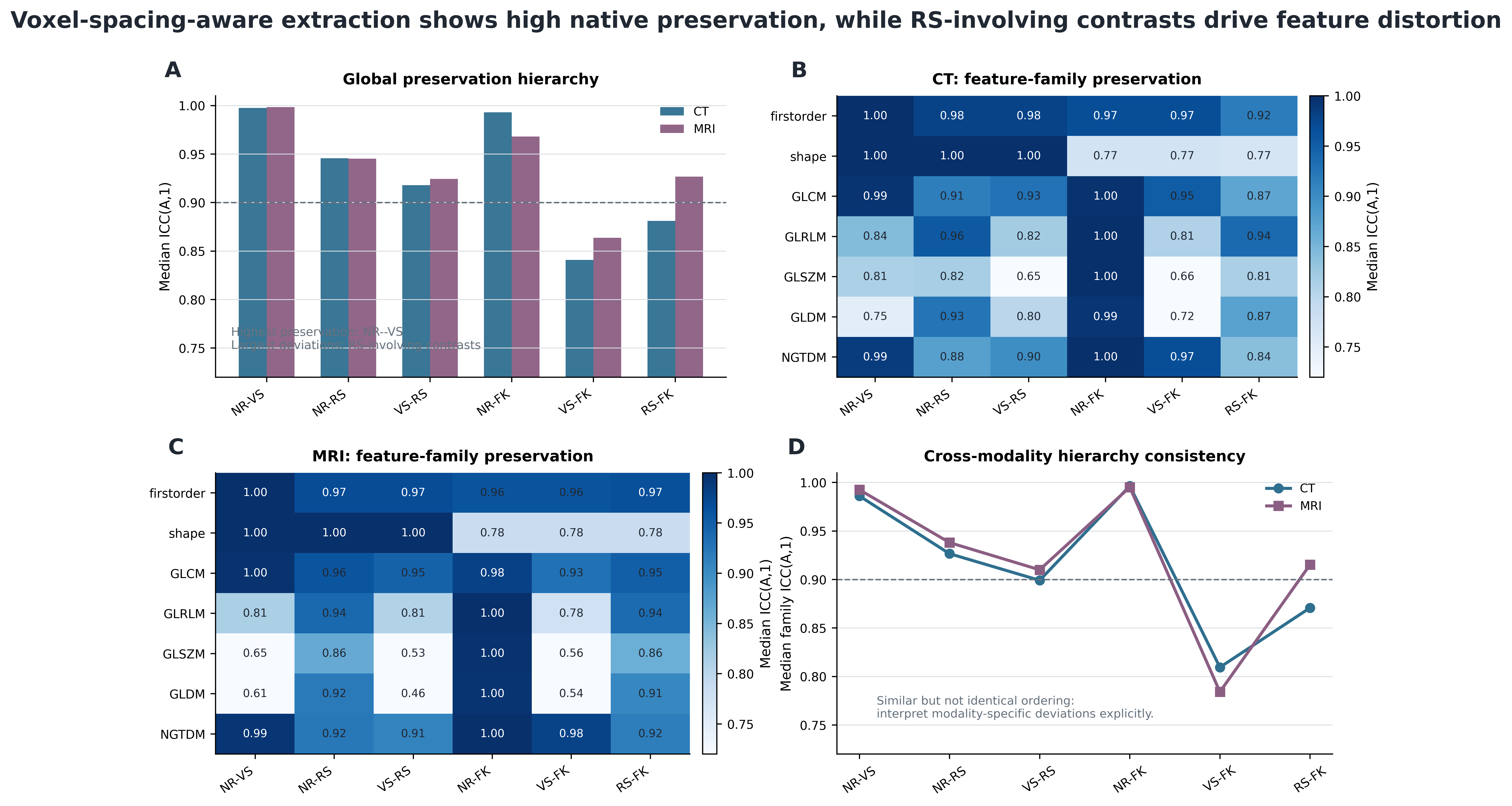}
\caption{
Radiomic preservation hierarchy across CT and MRI. Pairwise ICC(A,1) analyses demonstrated near-perfect preservation between NR and VS in both modalities, whereas comparisons involving RS showed larger deviations. Texture-derived descriptors demonstrated greater sensitivity to geometric preprocessing than global intensity-derived features. The same hierarchy was conserved across CT and MRI.
}
\label{fig:preservation_hierarchy_ct_mri}
\end{figure}

\subsection{Statistical Robustness Across Geometric Preprocessing}

A total of 196 radiomic descriptors derived from Original and Gradient-filtered images were evaluated across the four geometric preprocessing configurations.

Overall reproducibility remained high but was more heterogeneous across feature families after voxel-spacing–aware texture extraction. First-order descriptors showed the highest robustness in both CT and MRI, whereas several neighborhood-dependent texture families showed greater geometry sensitivity, particularly GLDM and GLSZM. GLCM and NGTDM showed comparatively preserved agreement in several pairwise comparisons, although their behavior remained dependent on modality and preprocessing contrast.

Across both modalities, NR–VS showed the highest pairwise preservation, with near-perfect median ICC(A,1) values and small absolute feature differences (Figure~\ref{fig:preservation_hierarchy_ct_mri}). However, this preservation was no longer uniform across all descriptors. In CT, 141/196 features showed ICC(A,1) $\geq 0.90$ for NR–VS comparisons, whereas MRI showed 140/196 features above this threshold. Features below this threshold were predominantly neighborhood-dependent texture descriptors, especially GLDM, GLRLM and GLSZM.

Comparisons involving isotropic resampling showed lower agreement and larger feature deviations, particularly for RS–VS and FK–RS (Table~\ref{tab:robustness_key_summary}). FK showed intermediate behavior despite preservation of the original voxel array, indicating that spacing reinterpretation alone altered a subset of geometry-sensitive descriptors.

Gradient-derived features showed lower robustness than Original-image features, with reduced ICC values and increased wCV and rMAD across preprocessing configurations.

The Friedman test identified significant preprocessing-related differences for most radiomic descriptors after FDR correction in both CT and MRI. Biological discrimination between benign and malignant lesions was broadly preserved across preprocessing strategies in CT, whereas MRI showed weaker discrimination overall.

\begin{landscape}

\begin{table}[htbp]
\centering
\scriptsize
\caption{
Key robustness patterns across geometric preprocessing strategies in CT and MRI.
Pairwise agreement metrics summarize global feature preservation between preprocessing configurations, whereas family-level metrics summarize overall robustness patterns across radiomic families.
}
\label{tab:robustness_key_summary}

\begin{tabular}{llcccc}
\hline
\textbf{Analysis} &
\textbf{Modality / family} &
\textbf{Key comparison} &
\textbf{ICC(A,1)} &
\textbf{ICC(C,1)} &
\textbf{Key variability metric} \\
\hline

\textbf{Native preservation}
& CT  & NR--VS & 0.998 & 0.998 & Median abs. diff. = 0.005 \\
& MRI & NR--VS & 0.998 & 0.999 & Median abs. diff. = 0.013 \\
\hline

\textbf{Effect of isotropic resampling}
& CT  & NR--RS & 0.946 & 0.953 & Median abs. diff. = 0.417 \\
& MRI & NR--RS & 0.945 & 0.967 & Median abs. diff. = 0.333 \\
\hline

\textbf{Largest absolute preprocessing deviation}
& CT  & FK--RS & 0.881 & 0.895 & Median abs. diff. = 2.175 \\
& MRI & FK--RS & 0.927 & 0.932 & Median abs. diff. = 1.004 \\
\hline

\textbf{Most robust family}
& CT  & First-order & 0.930 & 0.935 & Median rMAD = 0.008 \\
& MRI & First-order & 0.965 & 0.972 & Median rMAD = 0.014 \\
\hline

\textbf{More geometry-sensitive families}
& CT  & GLSZM / GLDM & 0.772 / 0.817 & 0.814 / 0.870 & rMAD = 0.057 / 0.083 \\
& MRI & GLDM / GLSZM & 0.659 / 0.772 & 0.816 / 0.881 & rMAD = 0.090 / 0.059 \\
\hline
\end{tabular}
\end{table}

\end{landscape}

\begin{figure}[htbp]
\centering
\includegraphics[width=\textwidth]{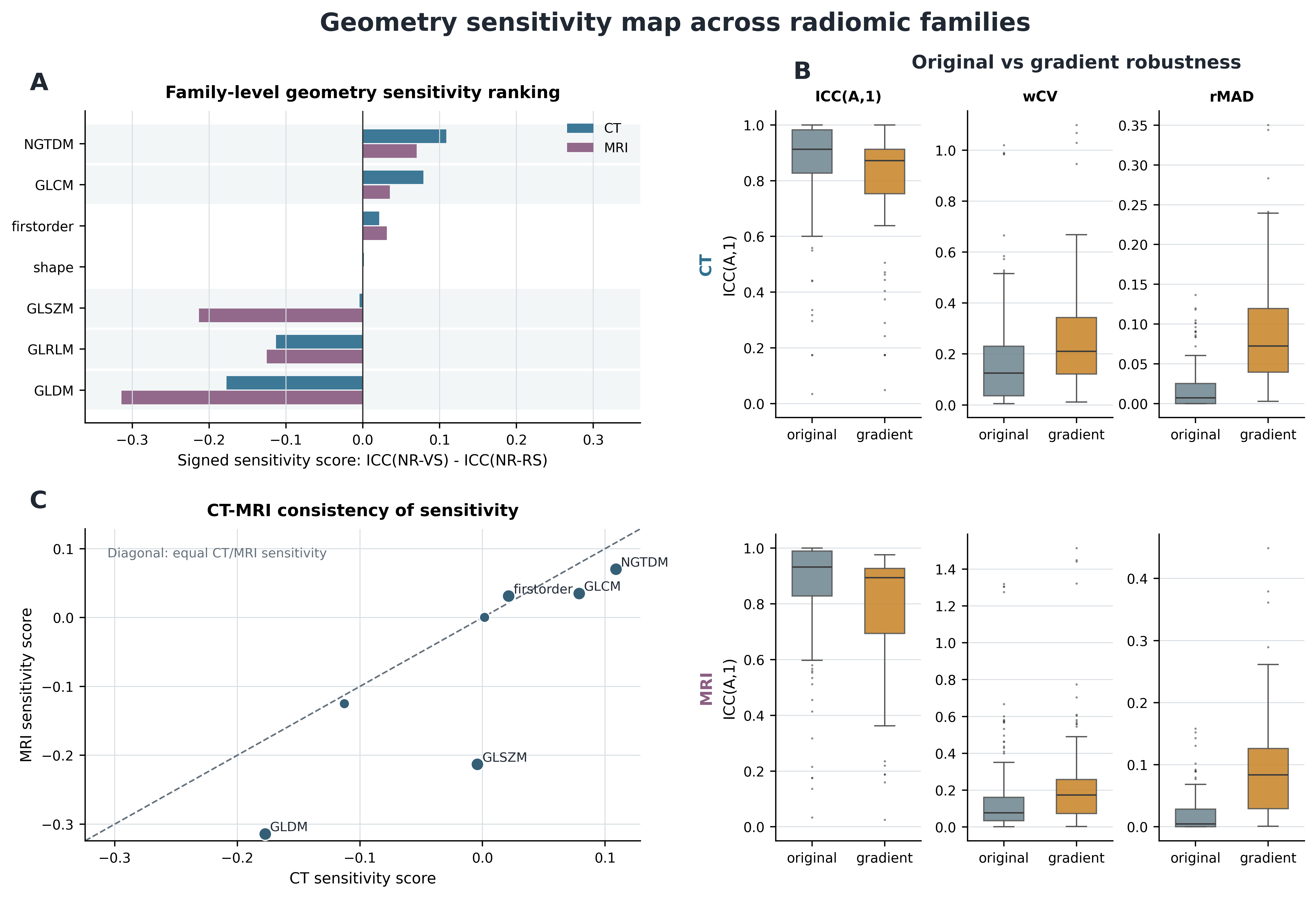}
\caption{
Geometry sensitivity across radiomic families and image representations in CT and MRI. Texture-derived and Gradient-based descriptors demonstrated greater sensitivity to geometric preprocessing than global intensity-derived features, with GLDM and GLSZM showing the greatest geometry dependence after voxel-spacing–aware texture extraction. Similar sensitivity hierarchies were observed across both modalities.
}
\label{fig:geometry_sensitivity_map}
\end{figure}

Geometry-related effects were not uniformly distributed across radiomic families. Neighborhood-dependent texture descriptors demonstrated substantially greater sensitivity to geometric preprocessing than global intensity-derived features, with the strongest effects observed for GLDM and GLSZM. Gradient-filtered representations consistently showed lower robustness and increased variability across both CT and MRI (Figure~\ref{fig:geometry_sensitivity_map}).

\subsection{Ensemble Feature Stability Analysis}

Ensemble-based feature-selection analysis demonstrated geometry-dependent differences in cross-fold feature stability.

Across both CT and MRI, isotropic resampling (RS) consistently produced the largest number of recurrent consensus descriptors under both relaxed and strict stability criteria. However, differences between preprocessing strategies were generally moderate, particularly under relaxed consensus settings, where all methods generated comparable numbers of stable features. Fake-isotropic preprocessing (FK) showed intermediate behavior despite preserving the original voxel array.

Stable ensemble descriptors were predominantly texture-derived and frequently enriched in Gradient-filtered representations. The most recurrent stable features originated mainly from GLDM, GLSZM and GLRLM families, with additional contributions from NGTDM and selected GLCM descriptors. This pattern was observed across both modalities and preprocessing strategies.

NR and VS demonstrated highly similar ensemble-selection behavior. Although voxel-spacing-aware extraction modifies several texture families through finite-volume and neighborhood-aware formulations, the resulting stable consensus signatures remained largely comparable between both configurations. This observation is consistent with the strong pairwise preservation previously observed between NR and VS and suggests that the most reproducible ensemble-selected descriptors are largely maintained after spacing-aware texture computation.

MRI reproduced the same overall preprocessing hierarchy despite substantially weaker biological discrimination than CT. Under strict consensus criteria, only a limited number of stable MRI descriptors remained, whereas relaxed consensus criteria identified recurrent Gradient-derived texture descriptors across all preprocessing strategies.

Overall, ensemble-selection behavior depended on geometric preprocessing strategy, particularly for texture-derived descriptors. Nevertheless, the substantial overlap observed between NR and VS suggests that voxel-spacing-aware extraction preserves much of the stable radiomic information identified by ensemble-based feature selection while avoiding interpolation-induced intensity modification.

\subsection{Predictive Modeling Analysis}

Predictive modeling analyses were performed using fully leakage-free pipelines, in which preprocessing, feature selection and hyperparameter optimization were learned exclusively from training data.

Under the common neutral signature setting, CT demonstrated highly similar predictive behavior for NR, VS and FK across Logistic Regression, XGBoost and MLP classifiers, whereas RS consistently showed lower held-out performance (Table~\ref{tab:ct_common_neutral_mlp}). Only two recurrent GLSZM-derived features remained shared across preprocessing configurations in CT.

\begin{table}[ht]
\centering
\caption{Common neutral signature modeling performance in CT including the MLP baseline.}
\label{tab:ct_common_neutral_mlp}
\begin{tabular}{llccc}
\hline
\textbf{Configuration} & \textbf{Model} & \textbf{Test ROC AUC} & \textbf{CV ROC AUC} & \textbf{Features} \\
\hline
NR & LR  & 0.763 & 0.722 & 2 \\
VS & LR  & 0.763 & 0.722 & 2 \\
FK & LR  & 0.763 & 0.722 & 2 \\
RS & LR  & 0.730 & 0.716 & 2 \\
\hline
NR & XGB & 0.763 & 0.690 & 2 \\
VS & XGB & 0.763 & 0.690 & 2 \\
FK & XGB & 0.763 & 0.690 & 2 \\
RS & XGB & 0.711 & 0.718 & 2 \\
\hline
NR & MLP & 0.782 & 0.716 & 2 \\
VS & MLP & 0.782 & 0.716 & 2 \\
FK & MLP & 0.782 & 0.716 & 2 \\
RS & MLP & 0.732 & 0.703 & 2 \\
\hline
\end{tabular}
\end{table}

Additional preprocessing-specific ensemble and unrestricted full-feature analyses reproduced the same overall CT pattern. Although RS generated larger and more recurrent texture-derived signatures, particularly enriched in Gradient-based descriptors, these changes did not consistently translate into superior held-out predictive performance. Across feature-selection settings and classifiers, NR, VS and FK generally matched or exceeded RS performance. Moreover, RS frequently exhibited larger train–test discrepancies, particularly in non-linear XGBoost models, suggesting that interpolation-induced feature regularization may facilitate optimization without necessarily improving generalization.

MRI demonstrated substantially greater dependence on geometric preprocessing strategy than CT. Unlike CT, the MRI common neutral signature retained 30 radiomic descriptors predominantly enriched in texture-derived and Gradient-filtered features. Under this setting, preprocessing effects became strongly classifier-dependent. VS achieved the highest held-out performance among Logistic Regression models (test ROC AUC = 0.700), outperforming NR (0.638), RS (0.599) and FK (0.574). In contrast, RS achieved the highest held-out performance among XGBoost models (test ROC AUC = 0.629), whereas VS showed markedly reduced performance in this setting (test ROC AUC = 0.380) (Table~\ref{tab:mri_common_neutral_modeling}). These findings indicate that the influence of geometric preprocessing depends not only on the extracted feature space but also on classifier complexity and inductive bias.

\begin{table}[ht]
\centering
\caption{Common neutral signature modeling performance in MRI.}
\label{tab:mri_common_neutral_modeling}
\begin{tabular}{llccccc}
\hline
\textbf{Configuration} & \textbf{Model} & \textbf{Test AUC} & \textbf{CV AUC} & \textbf{Train AUC} & \textbf{Sens.} & \textbf{Spec.} \\
\hline
NR & LR  & 0.638 & 0.587 & 0.670 & 0.643 & 0.548 \\
RS & LR  & 0.599 & 0.625 & 0.800 & 0.500 & 0.613 \\
VS & LR  & 0.700 & 0.575 & 0.625 & 0.714 & 0.645 \\
FK & LR  & 0.574 & 0.577 & 0.785 & 0.714 & 0.484 \\
\hline
NR & XGB & 0.558 & 0.557 & 0.975 & 0.357 & 0.645 \\
RS & XGB & 0.629 & 0.599 & 0.968 & 0.429 & 0.710 \\
VS & XGB & 0.380 & 0.532 & 0.984 & 0.286 & 0.645 \\
FK & XGB & 0.521 & 0.638 & 0.980 & 0.357 & 0.548 \\
\hline
\end{tabular}
\end{table}

Across preprocessing-specific ensemble, unrestricted full-feature and MLP analyses, no single preprocessing strategy consistently dominated across all classifiers and modeling settings. In CT, NR, VS and FK showed highly comparable performance and generally outperformed RS. In MRI, VS frequently achieved favorable performance in linear models and reduced-dimensionality settings, whereas RS occasionally achieved superior results in selected high-capacity XGBoost analyses. However, these RS advantages were not consistently reproduced across feature-selection strategies, modalities or classifier families. Overall, the results suggest that voxel-spacing–aware extraction preserves competitive predictive performance while avoiding interpolation-induced signal modification, supporting its use as a physically interpretable alternative to isotropic resampling.

\subsection{External Validation}

External validation was performed using independent private CT and MRI cohorts acquired under different clinical and technical conditions from the primary development datasets.

\begin{figure}[htbp]
\centering
\includegraphics[width=\textwidth]{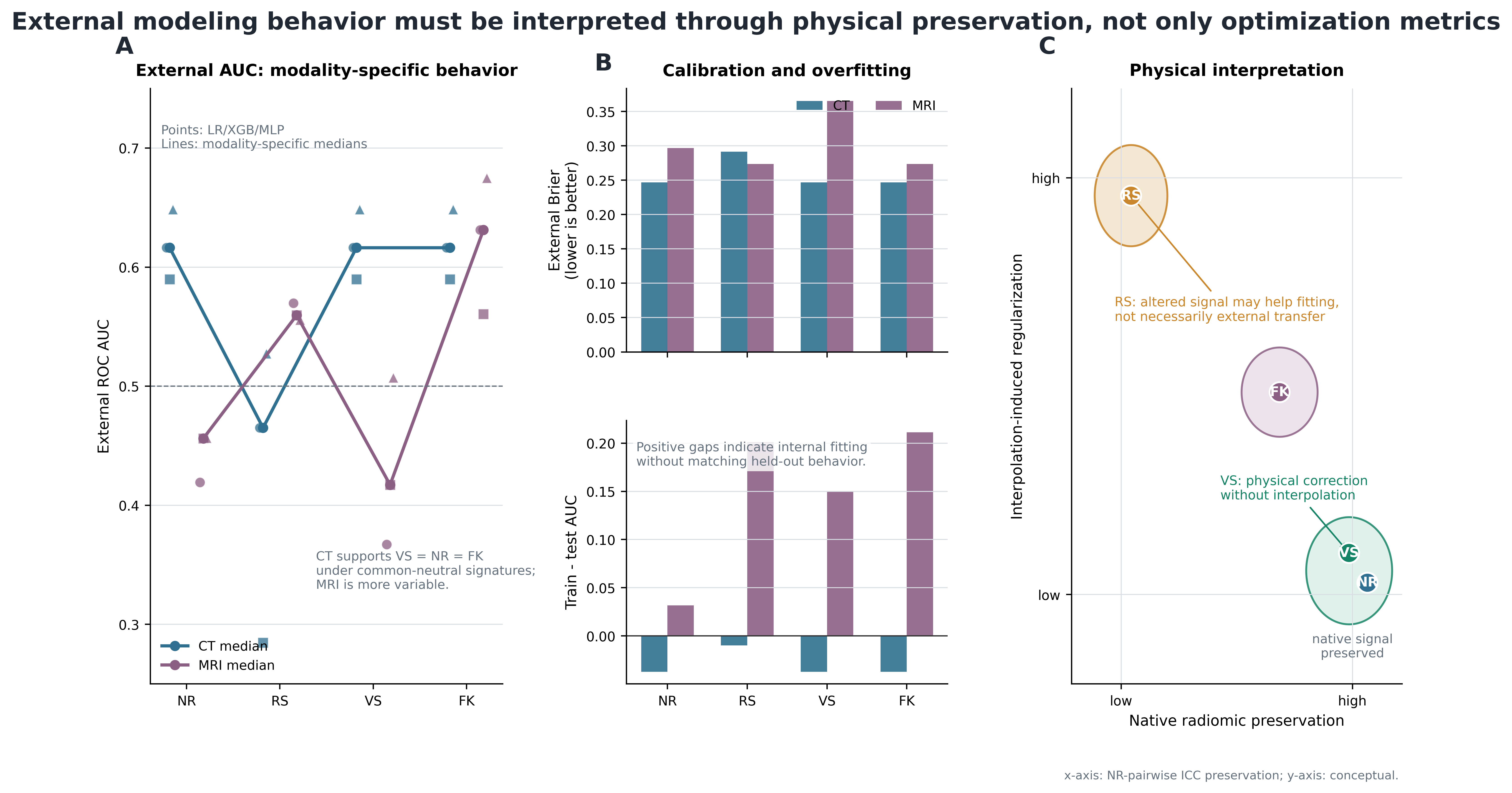}
\caption{
External modeling behavior across geometric preprocessing strategies. In CT, VS preserved predictive behavior identical to native extraction while RS consistently showed reduced discrimination and calibration. MRI demonstrated greater variability across preprocessing strategies, with no single approach consistently dominating across classifiers and validation settings.
}
\label{fig:modeling_generalization_summary}
\end{figure}

In CT external validation, VS preserved predictive performance identically to NR and FK across Logistic Regression, XGBoost and MLP models under the common neutral signature setting (Table~\ref{tab:ct_external_common_neutral}; Figure~\ref{fig:modeling_generalization_summary}). In contrast, RS consistently demonstrated lower external discrimination and poorer calibration behavior. The largest performance degradation was observed for XGBoost models, where RS achieved an external ROC AUC of only 0.284 compared with 0.590 for NR, VS and FK.

\begin{table}[ht]
\centering
\caption{External CT validation performance using the common neutral signature.}
\label{tab:ct_external_common_neutral}
\begin{tabular}{llccc}
\hline
\textbf{Configuration} & \textbf{Model} & \textbf{AUC} & \textbf{Brier} & \textbf{Specificity} \\
\hline
NR & LR  & 0.616 & 0.247 & 0.609 \\
VS & LR  & 0.616 & 0.247 & 0.609 \\
FK & LR  & 0.616 & 0.247 & 0.609 \\
RS & LR  & 0.465 & 0.292 & 0.087 \\
\hline
NR & XGB & 0.590 & 0.268 & 0.304 \\
VS & XGB & 0.590 & 0.268 & 0.304 \\
FK & XGB & 0.590 & 0.268 & 0.304 \\
RS & XGB & 0.284 & 0.357 & 0.174 \\
\hline
NR & MLP & 0.648 & 0.245 & 0.522 \\
VS & MLP & 0.648 & 0.245 & 0.522 \\
FK & MLP & 0.648 & 0.245 & 0.522 \\
RS & MLP & 0.527 & 0.256 & 0.261 \\
\hline
\end{tabular}
\end{table}

MRI external validation demonstrated substantially greater variability across preprocessing configurations and classifiers than CT. Unlike the internal validation analyses, VS did not maintain superior external performance. Across common-neutral validation settings, FK consistently achieved the highest discrimination, reaching ROC AUC values of 0.631, 0.561 and 0.675 for Logistic Regression, XGBoost and MLP models, respectively. RS generally showed intermediate performance, whereas VS frequently demonstrated the lowest external discrimination and calibration performance.

These findings suggest that the impact of geometric preprocessing on generalization is modality-dependent. In CT, voxel-spacing–aware extraction preserved predictive behavior equivalent to native extraction while avoiding interpolation-induced signal modification. In MRI, however, external generalization appeared to be influenced more strongly by dataset-specific characteristics and classifier interactions than by any single geometric preprocessing strategy.

\section{Discussion}

This study investigated the influence of voxel geometry handling on radiomic feature behavior across anisotropic CT and MRI datasets using a controlled multimodal framework. The principal finding was that voxel-spacing–aware extraction (VS) preserved native radiomic behavior with near-native agreement relative to non-resampled extraction (NR), while avoiding interpolation of the original image signal. In contrast, isotropic resampling (RS) produced larger alterations in feature distributions, particularly among texture- and Gradient-derived descriptors and generated more recurrent feature-selection patterns that did not consistently translate into improved predictive performance or external generalization.

These findings support the view that voxel anisotropy is not merely a technical nuisance but a fundamental determinant of neighborhood-dependent radiomic behavior. Previous studies have demonstrated that radiomic features are sensitive to acquisition parameters, reconstruction settings, voxel size and interpolation procedures \cite{mackin2015measuring,berenguer2018radiomics,zhao2016reproducibility,shafiq2017intrinsic,larue2017influence,whybra2019assessing}. However, conventional isotropic resampling simultaneously modifies voxel geometry and image intensities, making it difficult to distinguish geometric correction from interpolation-induced smoothing and regularization effects.

The proposed voxel-spacing–aware framework addresses this limitation by separating geometric modeling from signal modification. Rather than enforcing isotropic geometry through interpolation, VS preserves the native voxel array while incorporating true voxel dimensions directly into neighborhood-dependent computations. The fake-isotropic control (FK) further demonstrated that modifying spacing metadata alone altered several geometry-sensitive descriptors despite preserving the original voxel array. This finding suggests that part of the variability commonly attributed to interpolation may instead arise from implicit geometric assumptions embedded in neighborhood definitions, directional filtering operations and spatial connectivity relationships. At the same time, FK did not reproduce the full behavior observed after isotropic resampling, indicating that spacing reinterpretation and interpolation-induced smoothing affect radiomic descriptors through partially distinct mechanisms.

One of the most consistent observations across both modalities was the strong preservation between NR and VS. Although median agreement approached unity, this should not be interpreted as complete equivalence. A subset of geometry-sensitive texture descriptors, particularly within GLCM, GLDM and GLSZM families, exhibited lower pairwise agreement. This behavior is physically expected because these descriptors depend directly on local spatial connectivity and neighborhood definitions, which are explicitly modified in the spacing-aware implementation. Nevertheless, the overall preservation observed between NR and VS indicates that physical-space neighborhood correction can be introduced without substantially altering the native radiomic structure of the image.

The greater sensitivity observed for texture-derived and Gradient-based descriptors was also consistent with previous literature. Gradient representations inherently depend on local spatial-frequency content and directional neighborhood relationships and are therefore more susceptible to voxel anisotropy and interpolation effects than global intensity statistics \cite{shafiq2017intrinsic,larue2017influence,whybra2019assessing}. The present results extend these observations by demonstrating that even spacing reinterpretation alone can influence Gradient-derived radiomic behavior in the absence of interpolation.

Ensemble feature-selection analyses provided additional insight. RS consistently generated larger numbers of recurrent and apparently stable texture signatures, particularly among Gradient-derived descriptors. However, this increased recurrence did not consistently improve predictive performance or external generalization. This observation suggests that isotropic interpolation may introduce regularization effects that increase feature stability and recurrence without necessarily preserving informative biological heterogeneity. Conversely, NR and VS frequently produced sparser consensus signatures despite exhibiting strong pairwise preservation. Rather than indicating reduced robustness, this pattern may reflect preservation of native radiomic variability without interpolation-induced smoothing. Similar concerns regarding artificially inflated reproducibility after aggressive preprocessing have been raised previously in radiomics reproducibility studies \cite{zwanenburg2019radiomics,van2020radiomics,koccak2022key}.

Predictive modeling analyses were broadly consistent with these observations. In CT, VS preserved predictive behavior almost identically to NR across Logistic Regression, XGBoost and MLP models in both internal and external validation, whereas RS frequently demonstrated lower external discrimination and calibration stability. In MRI, the effect of preprocessing was more dependent on classifier architecture. RS occasionally achieved superior performance in selected XGBoost analyses, whereas VS often performed more favorably in lower-dimensional and linear-model settings. These findings suggest that part of the empirical benefit historically attributed to isotropic resampling may derive from interpolation-induced smoothing and feature-space regularization that facilitate optimization of flexible non-linear models rather than from improved preservation of native radiomic structure.

Despite substantial differences between CT and MRI image formation physics, the same geometric preservation hierarchy remained consistent across modalities. NR–VS comparisons consistently showed the strongest preservation, whereas analyses involving isotropic resampling demonstrated larger deviations. This consistency suggests that the observed effects arise primarily from spatial-neighborhood modeling and physical voxel relationships rather than from modality-specific intensity characteristics.

Taken together, these findings support voxel-spacing–aware extraction as a physically grounded alternative to isotropic resampling. The objective of VS is not necessarily to maximize predictive performance across all datasets and classifiers, but rather to provide a geometrically coherent framework that explicitly models anisotropic spatial relationships while minimizing modification of the original image signal. The present results demonstrate that this objective can be achieved while maintaining competitive predictive performance across both CT and MRI.

From a practical perspective, the proposed framework was implemented directly within the PyRadiomics backend while preserving standard behavior when spacing-aware extraction is disabled, facilitating integration into existing radiomic workflows.

Several limitations should be acknowledged. First, the implementation was evaluated within a modified PyRadiomics framework and requires independent validation in other software environments. Second, MRI demonstrated greater variability than CT across preprocessing strategies and classifiers, particularly during external validation. Third, only Original and Gradient-filtered representations were evaluated to preserve interpretability. Future studies should assess whether similar geometric effects extend to wavelet, Laplacian-of-Gaussian and other higher-order transformations. Fourth, no dedicated phantom validation or independent IBSI benchmarking of the spacing-aware implementation was performed. Finally, shape descriptors were intentionally excluded from the primary ensemble and predictive analyses to reduce geometry-dominated confounding.

Future work should investigate whether physically aware radiomic extraction improves multicenter reproducibility, harmonization robustness and deep-learning generalization in heterogeneous imaging environments. More broadly, the interaction between geometry-aware extraction, intensity normalization and harmonization strategies remains insufficiently understood and may play an important role in the development of reproducible quantitative imaging biomarkers. Because convolutional neural networks also rely on local spatial-neighborhood relationships, the implications of physically aware spatial modeling may extend beyond handcrafted radiomics to deep-learning-based image representations.

\section{Conclusion}

Voxel geometry substantially influences radiomic feature behavior in anisotropic CT and MRI acquisitions. While isotropic resampling introduced measurable alterations in texture structure, discretization behavior and feature-space organization, voxel-spacing–aware extraction preserved native radiomic behavior while explicitly incorporating physical voxel geometry during neighborhood-based computation without interpolation of the original image signal. Across both modalities, VS maintained competitive predictive performance while providing a physically coherent alternative to interpolation-based preprocessing. These findings support physically aware radiomic extraction as a promising strategy for improving the methodological coherence, interpretability and reproducibility of radiomics in anisotropic medical imaging.

\section*{Acknowledgments}
The authors thank the contributors of the LIDC-IDRI and I-SPY2/TCIA public datasets for making these imaging resources available for research.

\section*{Funding}
No specific funding was received for this study.

\section*{Conflict of Interest Statement}
The authors have no relevant conflicts of interest to disclose.

\section*{Ethics Statement}

The primary analyses were performed using publicly available de-identified datasets obtained from The Cancer Imaging Archive (TCIA).

The external validation cohorts were retrospectively collected under institutional ethics committee approval for imaging biomarker research using de-identified clinical imaging data. All procedures were conducted in accordance with institutional guidelines and the Declaration of Helsinki. The requirement for informed consent was waived due to the retrospective nature of the study and data anonymization.

\section*{Data Availability Statement}
The public datasets analyzed in this study are available from The Cancer Imaging Archive. Derived feature tables and analysis outputs may be made available by the corresponding author upon reasonable request, subject to institutional and data-sharing restrictions.

\section*{Code Availability Statement}
The modified PyRadiomics implementation and analysis scripts are being prepared for public release. Code may be made available upon reasonable request.

\bibliographystyle{unsrtnat}
\bibliography{biblex}

\end{document}